\documentclass[11pt,a4paper]{article}
\usepackage[hyperref]{emnlp2020}
\usepackage{times}
\usepackage{latexsym}

\usepackage{microtype}

\usepackage{subcaption}

\usepackage{graphicx}

\usepackage{amsmath}

\usepackage{url}
\usepackage{booktabs}
\usepackage{adjustbox}

\usepackage[final]{changes}

\aclfinalcopy % Uncomment this line for the final submission

\title{Interpreting Attention Models with Human Visual Attention\\ in Machine Reading Comprehension}

\author{Ekta Sood$^1$, Simon Tannert$^2$, Diego Frassinelli$^3$, Andreas Bulling$^1$, Ngoc Thang Vu$^2$\\
$^1$University of Stuttgart, Institute for Visualization and Interactive Systems (VIS), Germany\\ $^2$University of Stuttgart, Institute for Natural Language Processing (IMS), Germany\\ $^3$University of Konstanz, Department of Linguistics, Germany\\
\texttt{\{ekta.sood,andreas.bulling\}@vis.uni-stuttgart.de}\\ \texttt{\{simon.tannert,thang.vu\}@ims.uni-stuttgart.de}\\ \texttt{diego.frassinelli@uni-konstanz.de}
}

\date{}

\begin{document}
\maketitle
\begin{abstract}
While neural networks with attention mechanisms have achieved superior performance on many natural language processing tasks, it remains unclear to which extent learned attention resembles human visual attention.
In this paper, we propose a new method that leverages eye-tracking data to investigate the relationship between human visual attention and neural attention in machine reading comprehension. 
To this end, we introduce a novel 23 participant eye tracking dataset - MQA-RC, in which participants read movie plots and answered pre-defined questions. 
We compare state of the art networks based on long short-term memory (LSTM), convolutional neural models (CNN) and XLNet Transformer architectures. 
We find that higher similarity to human attention and performance significantly correlates to the LSTM and CNN models. However, we show this relationship 
does not hold true for the XLNet models -- despite the fact that the XLNet performs best on this challenging task.
Our results suggest that different architectures seem to learn rather different neural attention strategies and similarity of neural to human attention does not guarantee best performance.
\end{abstract}

%%%%%%%%%%%%%%%%%%%%%%%%%%%%%%%%%%%%%%%%%%%%%%%%%%%%%%%%%%%%%%%%%%%

\section{Introduction}
Due to the high ambiguity of natural language, humans have to detect the most salient information in a given text and allocate a higher level of attention to specific regions to successfully process and comprehend it~\cite{schneider1977controlled,shiffrin1977controlled,poesio1995semantic}.
Eye tracking studies have been extensively used in various reading comprehension tasks to capture and investigate these attentive strategies~\cite{rayner2009eye} and have, as such, helped to interpret cognitive processes and behaviors during reading.

Attention mechanisms in neural networks have been inspired by human visual attention \cite{Bahdanau:2014vz, hassabis2017neuroscience}.
Similar to humans, they allow networks to focus and allocate more weight to different parts of the input sequence~\citep{mnih2014recurrent,chorowski2015attention,xu2015show,vaswani2017attention,jain2019attention}.
As such, neural attention can be viewed as a model of visual saliency that makes predictions over the elements in the network's input -- whether a region in an image or a word in a sentence~\cite{frintrop2010computational}.
Attention mechanisms have recently gained significant popularity and have boosted performance in natural language processing tasks and computer vision~\cite{ma2003contrast,sun2003object,seo2016bidirectional,velivckovic2017graph,sood20_neurips}.

Although attention mechanisms can significantly improve performance for different NLP tasks, performance degrades when models are exposed to inherent properties of natural language, such as semantic ambiguity, inferring information, or out of domain data~\cite{blohm2018comparing,niven2019probing}. These findings encourage work towards enhancing network's generalizability, deterring reliance on the closed-world assumption~\cite{reiter1981closed}.
In machine reading comprehension (MRC), it has been proposed that the more similar systems are to human behavior, the more suitable they become for such a task~\citep{trischler2016newsqa,luo-etal-2019-reading,zheng2019human}.
As a result, much recent work aims to build machines which read and understand text, mimicking specific aspects of human behavior 
~\citep{hermann2015teaching,nguyen2016ms,rajpurkar2016squad,blohm2018comparing}.
To that end, by employing self-attention, researchers attempt to enhance comprehension by building models which better capture deep contextual and salient information~\cite{vaswani2017attention,devlin2018bert, shen2018disan,yu2018qanet,zhang2018self}.

\begin{figure}[t]
  \centering
  \includegraphics[width=1.1\linewidth]{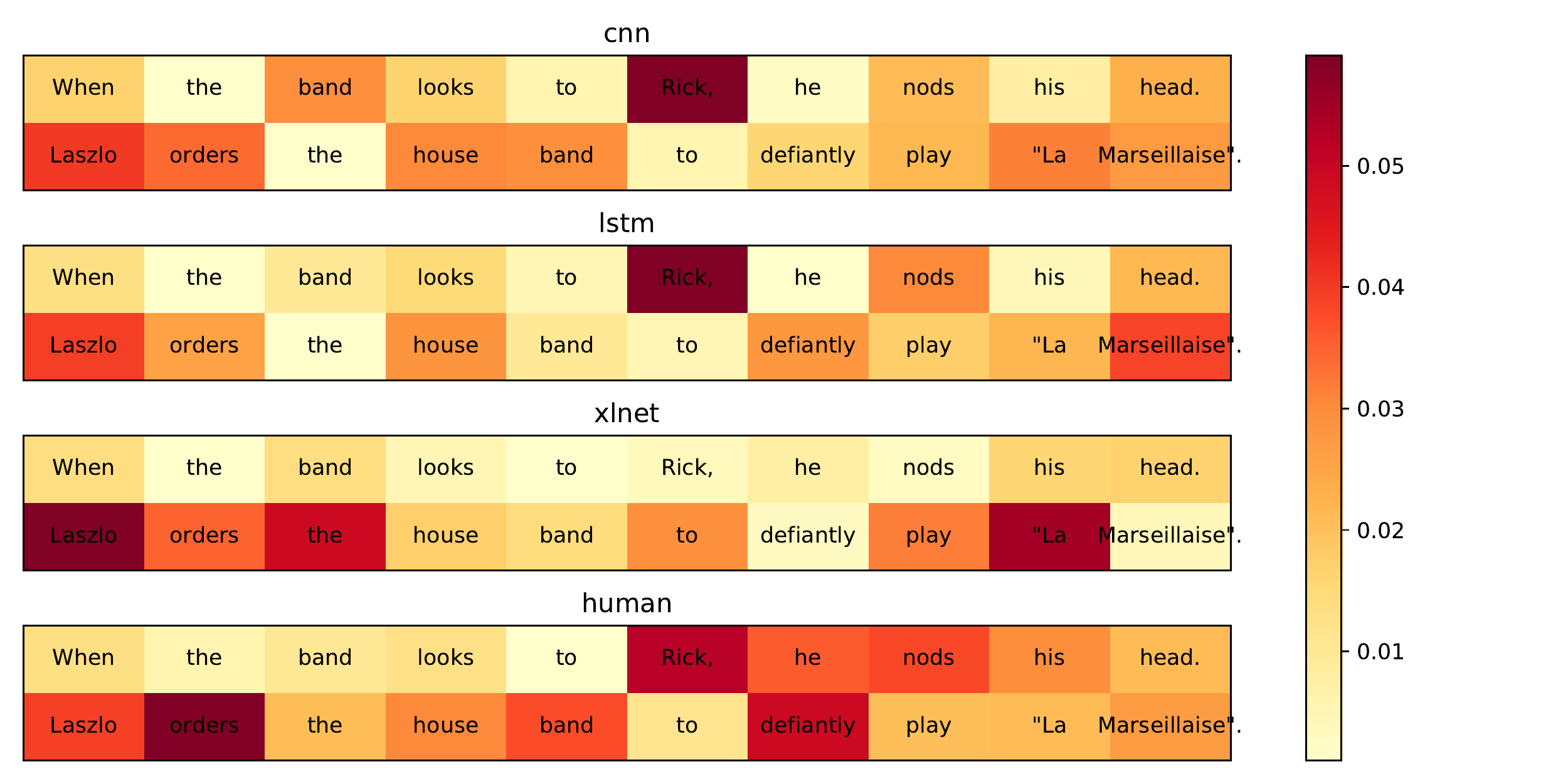}
  \caption{Example attention distributions of neural models (cnn, lstm, xlnet) and humans.}
 \label{fig:attention_heat}
\end{figure}

\begin{figure}[t]
  \centering
  \includegraphics[width=\linewidth]{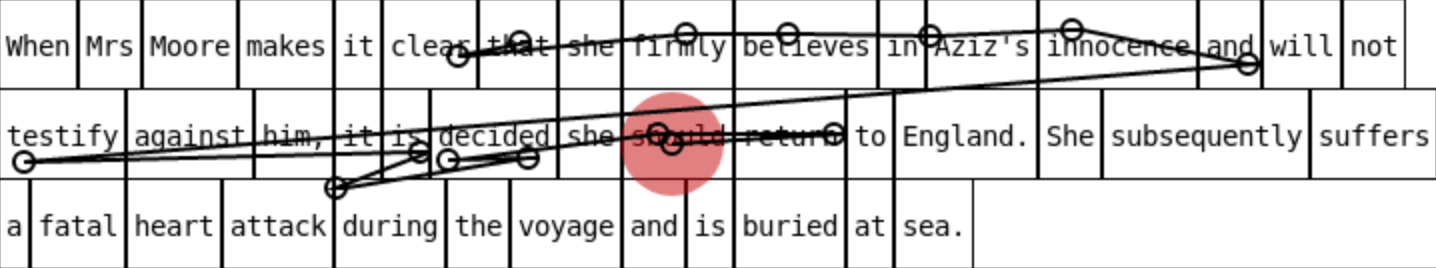}
 \caption{An exemplary scan path shows a reading pattern. The red circle corresponds to the location of the current fixation. Its size is proportional to the duration of the fixation.}
 \label{fig:attention_scanpath}
\end{figure}

As neural attention allows us to ``peek'' inside neural networks, it can help us to better understand how models make predictions (see Figure~\ref{fig:attention_heat}). 
Similarly, human visual attention (which is captured by physiological data such as eye tracking), allows us to quantify the relative importance of items within the visual field when reading texts (see Figure \ref{fig:attention_scanpath}).

In this work, we propose a novel method that leverages human eye tracking data to investigate the relationship between neural performance and human attention strategies.
Concretely, by interpreting and comparing the relationship between neural attention distributions of three state of the art MRC models to human visual attention, our research for the first time addresses the following questions: (i) What is the correlation between a particular network behavior and the human visual attention?
(ii) Is the emulation of the human attention system the reason why neural models with attention mechanisms achieve state of the art results on machine reading comprehension tasks? 

To answer these questions, we first extend 
the MovieQA dataset~\cite{MQA} with eye tracking data. In addition, we present a novel visualization tool to qualitatively compare the differences in attentive behaviors between neural models and humans by showing their patterns over time in a split screen mode.
Second, as widely suggested in the cognitive science literature, we quantify human attention in terms of the word-level gaze duration recorded in our eye tracking dataset~\cite{rouse1986looking,milosavljevic2008first,van2012intentional,lipton2016mythos,wiegreffe2019attention}. 
Third, we interpret the relationship between human attention and three 
state of the art systems 
based on CNN, LSTM, and XLNet~\cite{HochreiterLSTM_97,yang2019xlnet} 
using Kullback-Leibler divergence~\cite{kullback1951information}.
By doing so, we are able to compare, evaluate and better understand neural attention distributions on text across these attention models. To the best of our knowledge, we are the first to propose a systematic approach for comparing neural attention to human gaze data in machine reading comprehension.

The main findings of our work are two-fold:
First, we show that there is a statistically significant correlation between the CNNs and LSTMs model performances and similarity to human attention. Second, we show that the behavior of LSTM models is significantly more similar to humans than the XLNet ones even though the latter perform best on the MovieQA dataset.

%%%%%%%%%%%%%%%%%%%%%%%%%%%%%%%%%%%%%%%%%%%%%%%%%%%%%%%%%%%%%%%%%%%

\section{Related Work}

\subsection{Eye-tracking for Attention and Comprehension}

Eye tracking studies have been extensively used in cognitive science research to investigate human attention over time \cite{rayner1998eye,wojciulik1998covert,tsai2012visual,eckstein2017beyond}.
Importantly, it has been demonstrated that attention and saccadic movements are strongly intertwined~\citep{hoffman1995role,deubel2000attention,kristjansson2011intriguing}.
Eye movement behaviors which are evoked from intricate information processing tasks, such as reading, can be used to identify visual attentional allocation~\citep{posner1980attention,posner1980orienting,henderson1992visual}.

As indicated in the \textit{Reading Model}~\cite{just1980theory}, we assume a strong relationship between eye fixations, attention, and reading comprehension. 
In their eye tracking study, ~\citet{just1980theory} measured cognitive processing load using fixation duration. Specifically, they found that participants look longer or more often at items that are cognitively more complex, in order to successfully process them. Cognitive load increases when readers are ``accessing infrequent words, integrating information from important clauses and making inferences at the ends of sentences''. 

\subsection{Attention Mechanisms}

In the attention-based encoder-decoder architecture, rather than ignoring the internal encoder states, the attention mechanism takes advantage of these weights to generate a context vector, which is used by the decoder at various time steps~\cite{Bahdanau:2014vz,Attention_Luong15,chorowski2015attention,CAM,HierarchicalDocClass_Yang16,dzendzik2017framed}.

In Transformer networks, the main differences to previous attentive models are that these networks are purely based on attention where LSTM or GRU units are not used, and attention is applied via self-attention and multi-headed attention~\cite{vaswani2017attention} without any order constraint. 
Since the introduction of pre-trained Transformer  networks, we have observed, on the one hand, a rise in state of the art performance across a multitude of tasks in NLP~\cite{devlin2018bert,radford2018improving,yang2019xlnet}.
On the other hand, much effort is needed to interpret these highly complex models (e.g. in ~\citet{vig2019visualizing}). 

\subsection{Question Answering and Machine Comprehension}

We use question answering (QA) tasks to compare human and machine attention.
Although such tasks have been widely explored with neural attention models, creating systems to comprehend semantically diverse text documents and answer related questions remains challenging~\cite{qiu2019survey}. These models tend to fail when faced with adversarial attacks: the type of noise humans can easily resolve~\cite{jia2017adversarial,blohm2018comparing,yuan2019adversarial}. These studies uncovered the limitations of QA systems, indicating that models might process text in a different manner than humans: they rely on pattern matching in lieu of human-like decision making processes which are required in comprehension tasks~\cite{just1980theory,posner1980attention,blohm2018comparing}. 

\subsubsection{Eye Tracking and Neural Networks} 

In the past years, researchers have started leveraging human gaze data for attentive neural modeling tasks.
For example,~\newcite{hahn2016modeling,hahn2018modeling} presented a neural QA network that combined both a task and attention module to predict and simulate human reading strategies.
The authors proposed the \textit{trade-off hypothesis:} human reading behaviors are task-specific and therefore evoke various specific strategies for each of these tasks.
To validate their hypothesis, they used eye tracking data as the gold standard and compare model predictions of zero or one (fixated or not).
In another work,~\newcite{das2017human} investigated the differences between neural and human attention over image regions in a visual question answering task.
Their method focused on correlation ranking and visualizations. 
Note that comparisons of human and neural attention distributions over text have not been explored so far. 
When the goal is to purely improve performance, several papers proposed integrating gaze data into neural attention as an additional variable in the equation or as a regularization method ~\cite{sugano2016seeing,barrett2018sequence,qiao2018exploring,sood20_neurips}.
\subsection{Neural Interpretability}

In order to further understand the behavior 
of neural networks, research in neural interpretability has grown dramatically in the recent years~\cite{lipton2016mythos,gilpin2018explaining,hooker2019benchmark}. Such methods include: introducing adversarial examples, error class analysis, modeling techniques (e.g. self-explaining networks), and post-hoc analysis of attention distributions~\cite{lipton2016mythos,alvarez2018towards,rudin2019stop,sen-etal-2020-human}.

To shed light on the decisions taken by these networks, multiple interpretability studies have investigated their outputs and predictions ~\cite{alvarez2018towards,blohm2018comparing,gilpin2018explaining},
and analyzed their behavior through loss visualization from various architectures~\cite{ribeiro2016should}.

Nevertheless, a real understanding of the internal processes of these black boxes is still rather limited~\cite{gilpin2018explaining}. Although these interpretations might explain predictions, there is still a lack of explanation
regarding the mechanisms by which models work as well as limited insight regarding the relationship between machine and human visual attention~\cite{lipton2016mythos}.

%%%%%%%%%%%%%%%%%%%%%%%%%%%%%%%%%%%%%%%%%%%%%%%%%%%%%%%%%%%%%%%%%%%

\section{Resources}
\subsection{MovieQA Dataset}
The MovieQA dataset~\citep{MQA} is used in all experiments conducted in this work.
The dataset was comprised of a variety of available sources, however for the tasks in this work we only use the plot synopses. 
The plots vary between 1 to 20 paragraphs in size, and are checked by annotators to ensure they consist of movie relevant events and character relationships. There are a total of almost 15,000 human generated questions in this dataset corresponding to 408 movie plots. Of the 5 answer candidates denoted for each question, there is only one with a correct answer and the rest are deceptive incorrect answers.
The data used for training all our models consists of plots with their corresponding questions: 9,848 training, 1,958 development and 3,138 test questions, respectively.

\subsection{Reading Comprehension with Eye Tracking Dataset} 
We present a novel reading comprehension eye tracking dataset\footnote{The dataset is available at \url{https://perceptualui.org/publications/sood20\_conll/}} - MQA-RC -
which allows researchers to observe changes in reading behavior in three comprehension tasks and to potentially induce processing strategies evoked by humans.
This new extension provides a gold standard to compare and synchronize model versus human visual attention in comprehension tasks.
To the best of our knowledge there are no available eye tracking datasets which use machine learning corpora as stimuli. Therefore, we build and use our reading comprehension gaze dataset as the gold standard. 
In addition, we provide coreference chains labeled by two human annotators\footnote{See appendix material for further information on coreference annotation}. 
Based on the lower fixation durations observed in the eye tracking data, we find that humans can easily resolve pronouns in the MQA-RC dataset (cf. Figure \ref{fig:saliency_coref}), where fixation durations are used to measure information processing load~\cite{arnold2000rapid,rahman2012resolving,EmrahCinkara}.
The figure also shows saliency over the proper nouns compared to their mentions in the chains.

\paragraph{Data collection}
Our dataset is based on two studies: in Study~1 we randomly selected a set of 16 documents on which the majority 
of both LSTMs and CNNs models failed to correctly answer the questions; in Study~2 we selected a different set of 16 documents on which the majority of models succeeded in predicting the correct answers.

In total, our dataset contains gaze data from 23 English native speakers who were recorded while reading 32 documents (around 200-250 words each) in three different comprehension tasks.
We used a Tobii 600Hz head-mount eye-tracker. In total, each session lasted 45 minutes including the time required for calibration and 5-minutes breaks every 15 minutes. 

\paragraph{Study 1}
For each of the 16 documents we designed three experimental conditions: 1) regular QA where the participants have access to the plot, the question, and five answer candidates; 2) open-ended answer generation where the participants see the plot and the question but have to generate their own responses; and 3) QA by memory where the participants can first read the plot and then answer to the question (5 possible answers) without having the plot available. In condition 3, participants have to recover information from memory in order to answer the question. To guarantee a balanced design, we divided the 48 experimental items in three schemes containing each document only once: 5-5-6 items (for condition 1-2-3) in schema A, 5-6-5 in schema B, and 6-5-5 in schema C. We randomly assigned each participant to one of these schemes where the order of the conditions followed a Latin Squared Design~\cite{bradley1958complete}.

\paragraph{Study 2}
We conducted a follow up study in which we took only the plots for which the majority of CNN and LSTM models predicted correctly. We hypothesized that such items that are, on average, easier for the models are also easier for the humans (higher correlation score).
In this study, we only collected data for the regular QA task (condition 1). The experiment was performed by five new participants. Each participant saw all the 16 plots in a randomized order. 
\footnote{In order to maintain the same amount of data samples for both study 1 and 2, we randomly selected a subset participants data from study 1. Instead of using the full 18 participants from study 1, we used 15 participants.}

\paragraph{Data analysis}
Table~\ref{tab:condition1_pts} shows the distribution of data, inter-annotator agreement, and accuracy observed on our MQA-RC dataset.
We show across both studies that humans agree on selected answers for the given questions  and are highly accurate.
It is important to note that we only use data from the regular QA task (condition 1) so that we can compare attention and performance for difficult vs. easy cases.

\begin{table}
  \adjustbox{max width=\columnwidth}{
  \begin{tabular}{r r r r r r} 
    \toprule
    Study & Schema & No. Doc & No. Participants & IAA & Acc
 \\ [0.5ex] 
    \midrule
    Study1 & A & 5 & 1-6 & 83.3\% & 93\%
    \\ 
    
    Study1 & B & 5 & 6-12 & 100\% & 100\%
    \\
    
    Study1 & C & 6 & 12-18 & 100\% & 100\%
    \\
    
    Study2 & No-Schema & 16 & 5 & 89.0\% & 95\%
    \\
    \bottomrule
  \end{tabular}
  }
  \caption{Distribution in MovieQA with eye tracking. We show the two different studies and the number of documents seen in each schema iteration. For study 1, there are three schema iterations (A, B, C) and for study 2 there are no schema iterations (as this is only for answer by selection). We also show the number of participants for each schema iteration, and the corresponding inter-annotator agreement (agreement on answer selected). Lastly, we show the accuracy of the participants for correctly answering each question in the respective study and schema iteration.}
  \label{tab:condition1_pts}
\end{table}

\paragraph{Visualization tool}
We developed a web interface tool\footnote{The tool is also available at \url{https://perceptualui.org/publications/sood20\_conll/}} to visualize the eye tracking data (cf. Figure \ref{fig:interface}).
This tool is simple, easy to use and can visualize any eye tracking data where text is used as the stimulus (see an example in Figure~\ref{fig:attention_scanpath}).
Inputs to the tool are two files -- one with eye tracking data and another with the corresponding text stimulus.
The eye tracking data consists of the $x$ and $y$ on-screen gaze coordinates, fixation duration for each word, and word IDs (cf. Figure \ref{fig:example_data}).
Our tool then maps the coordinates to the stimulus and provides real time scanpaths visualization.
In addition, our tool can compare neural and human visual attention
via linear visualization (left to right) with a split screen (e.g., left side model, right side human).
This functionality allows users to observe, in real time, the dynamic network and human visual attention distributions.

%%%%%%%%%%%%%%%%%%%%%%%%%%%%%%%%%%%%%%%%%%%%%%%%%%%%%%%%%%%%%%%%%%%

\section{Neural Models} 
\subsection{Two Staged Attention Models} 
We re-implement both the CNN and LSTM QA ensemble models with two staged attention from~\newcite{blohm2018comparing} that provides state of the art results on the MovieQA dataset~\cite{MQA}.
This is a multiple choice QA task in which each datapoint contains the plot of a movie as well as its corresponding question and five potential answer candidates. 
The models are based on the compare-aggregate framework. 
Concretely, the models compare the plot to the respective question and aggregates this comparison into one vector representation to obtain a confidence score after applying the softmax, for each answer candidate. 
The best results were obtained from the majority vote of the nine best performing models.

The two-staged attention is performed at the word and at sentence level, where the plot is weighted with respect to the question or a possible answer candidate.

\begin{align}
G &= \text{softmax}\left (X^{T}P \right)\\
H &= XG
\end{align}

The word level $X$ indicates the answer candidate (5 total) or the question. Subsequently, when computing sentence level attention, the question or answer candidate are represented as such.~\newcite{blohm2018comparing} apply the dot-product computation for the attention mechanism. The two variations of this model with CNN and LSTM models provided state of the art results on the MovieQA dataset with an average of 84.5\% on the validation set and an average of 85\% on the test set.

The authors performed a case study to further investigate the comprehension limitations of the models compared to human inference.
In their analysis, they compared both networks against human performance in order to infer processing strategies which human possess but are not shown by the models. They investigated the most difficult cases, where the majority of both nine best models failed to correctly answer the question.
This motivates why we used the difficult and easy documents for the CNN and LSTM models~\cite{blohm2018comparing}, as they are the only paper to date which both obtain SOTA results and offered qualitative analysis on the gap between human and model performance.
When the majority of the models fail to correctly answer the question, we classify these documents as \textit{difficult} cases for the two networks; vice versa for the \textit{easy} documents.

\subsection{XLNet Models}

We used the pre-trained XLNet model and fine-tuned it for the QA task~\cite{MQA,yang2019xlnet}.
We opted for XLNet given that it is a recent Transformer network for language understanding that outperformed BERT and other large-scale pre-trained language models on a variety of NLP tasks~\cite{yang2019xlnet}.
It was trained on large corpora with training objectives which are compatible with unsupervised learning and can be fine-tuned to new tasks and datasets. 

XLNet is based on an auto-regressive approach in which the model uses observations from previous time steps in order to predict the weight for the next time step. Advancing from the traditional auto-regressive approach, such as a Bidirectional LSTM, the authors also combine their network with an auto-encoding approach seen with the BERT model~\cite{devlin2018bert}. By combining both approaches, XLNet introduces permutations on both sides. Moreover, the self-attention network~\cite{vaswani2017attention} uses three components, queries, keys and values, all of which are calculated from their respective embeddings. The output is a weighted sum of the values, in which the values are weighted with a score calculated as the dot product of the respective queries and keys. It is important to note that the queries are related to the output and the keys are related to the given input. 
During fine-tuning, however, the model is essentially the Transformer-XL~\cite{vaswani2017attention,dai2019transformer,yang2019xlnet}. The auto-regressive language model estimates the joint probability over the input elements (in XLNet this $x$ is language agnostic, i.e it is a subtoken). 

\begin{equation}
  P(X) = \prod_t P(x_t | X_{<t})
\end{equation}

The input sequence is the concatenation of each $x$ in the plot with the question and a potential answer candidate (there are five possible answer candidates and one correct answer).

When fine-tuning on the question answering task, the model objective is multi-label classification given an input sequence. Note, the permutation language model is the component which helps XLNet capture longer dependencies between elements in a given input sequence~\cite{yang2019xlnet}. In our method, we fine-tune the XLNet with 24 attention layers and 16 attention heads \cite{yang2019xlnet}.
The fine-tuned model makes a prediction by applying the argmax over the softmax, selecting the potential y-label, or answer candidate, with the highest confidence scores. The fine-tuned XLNet outperforms all other results on the validation set, obtaining the new highest accuracy of 91\%.

%%%%%%%%%%%%%%%%%%%%%%%%%%%%%%%%%%%%%%%%%%%%%%%%%%%%%%%%%%%%%%%%%%%

\section{Analysis Method}

%%%%%%%%%%%%%%%%%%%%%%%%%%%%%%%%%%%%%%

\subsection{Human Gaze-Attention Extraction}
We obtain token level gaze counts (frequency counts) by mapping the $x,y$ coordinates to bounding boxes set around each word of the stimuli. We convert the raw gaze counts into a probability distribution over the document by dividing each gaze count by the sum of all gaze counts. These token level frequency counts obtained in the hit testing method, reflecting gaze duration: the more often a token of the text is attended to, the more important it is for humans to answer the question~\citep{just1980theory}. 

We extract word level attention weights and average them over documents, thereby comparing the word attention at document level. Since for humans, the task is to read the entire short document and then answer the question given the entire context, all items within the context are interconnected. Therefore, it is misleading to only analyze attention over one sentence or one part of the document. Furthermore, it is not cognitively plausible to limit comparison to attention distribution over specific sentences or only part of the documents.

\subsection{Extracting LSTM and CNN Word Level Attention}
The sentence level attention for the CNN and LSTM models have very low entropy, where essentially almost all of the attention is distributed to one sentence and the rest of the sentence attention weights are almost zero. This is a property of the two-staged attention, which XLNet does not have.
Therefore, we leverage word level attention to compare model attention versus human visual attention.
During evaluation, we extract token attention weights for each of the nine best models. We then ensemble the neural attention weights. Figure~\ref{fig:CNN_entropy_word} and~\ref{fig:LSTM_entropy_word} in the Appendix show the word level attention distribution of CNN and LSTM models.

\subsection{Extracting XLNet Word Level Attention}

We extracted the attention weights from the nine best XLNet models by leveraging the output of the last attention layer. It contains token level weights for each plot-answer candidate pairing. More specifically, the output of the last attention layer is a matrix of 1024 x 1024, which contains a vector of attention weights vectors for each respective token.
We did so because in Transformers, attention computations happen simultaneously, while for LSTMs and CNNs they happen last. In order to compare XLNet to the LSTM and CNN models, we therefore only take the final output of the self-attention layer.
Furthermore, to make these weights comparable to human gaze attention we take the maximum value in each token vector~\cite{htut2019attention} and normalize them by the sum of the weights.

\subsection{Attention Comparison Metrics}
\paragraph{KL divergence}
In order to compare the human and neural attention distributions, we computed the Kullback-Leibler divergence~\cite{kullback1951information}. Concretely in this paper, we calculate the KL divergence for average-human to average-model along the word level attention distributions. 
This method is used to compare two probability distributions, akin to relative entropy. The output will reflect an understanding of the differences between the two probability distributions (cf. Equation~\ref{eq:KL_div}).

\begin{equation}
\label{eq:KL_div}
{\displaystyle D_{\text{KL}}(H\parallel M)=\sum _{x\in {\mathcal {X}}}P(x)\log \left({\frac {H(x)}{M(x)}}\right).}
\end{equation}

where H stands for the human attention distribution and M for the model attention distribution.

\paragraph{Spearman's rank correlation}
Spearman's rank correlation coefficient is used to discover the relationship between two variables \citep{zar1972significance}. We use the standard Spearman's rank correlation coefficients implementation from SciKit-Learn~\cite{kokoska2000crc,scikit-learn}, to measure if there is a correlation between model performance and the KL divergence between models and humans attention distributions. Model performance refers to the number of models that provide correct answers in the ensemble setting.
Because KL divergence reflects the differences between distribution, i.e. lower divergence means high similarity to human visual attention, a negative Spearman's rank correlation indicates that higher performance means high similarity to human visual attention. 
The p-value indicates the significance and the likelihood that the null hypothesis will be rejected. With p-values below 0.01, we can reject the null hypothesis and thus accept that there is a statistically significant correlation between divergence and accuracy.

%%%%%%%%%%%%%%%%%%%%%%%%%%%%%%%%%%%%%%%%%%%%%%%%%%%%%%%%%%%%%%%%%%%

\section{Analysis Results}
\label{sec:Analysis Results}

\subsection{Models vs. Humans}

\begin{figure*}[h]
  \centering
  \begin{subfigure}{.48\textwidth}
    \centering
    \includegraphics[width=\linewidth]{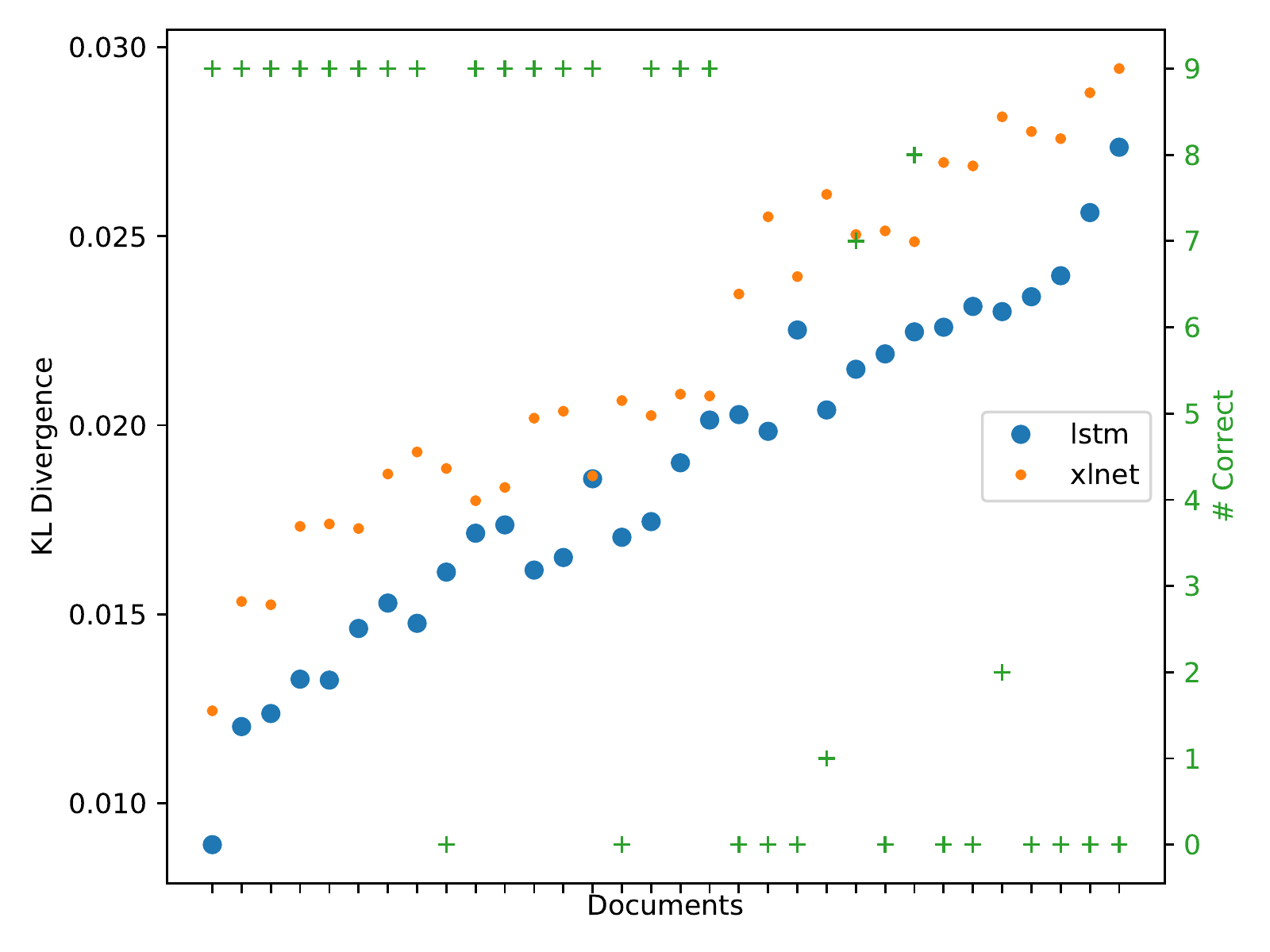}
    \caption{LSTM versus humans --- KL divergence and number of correct models per document.}
    \label{fig:LSTM_KL_Correctness}
  \end{subfigure}
  \hfill
  \begin{subfigure}{.48\textwidth}
    \centering
    \includegraphics[width=\linewidth]{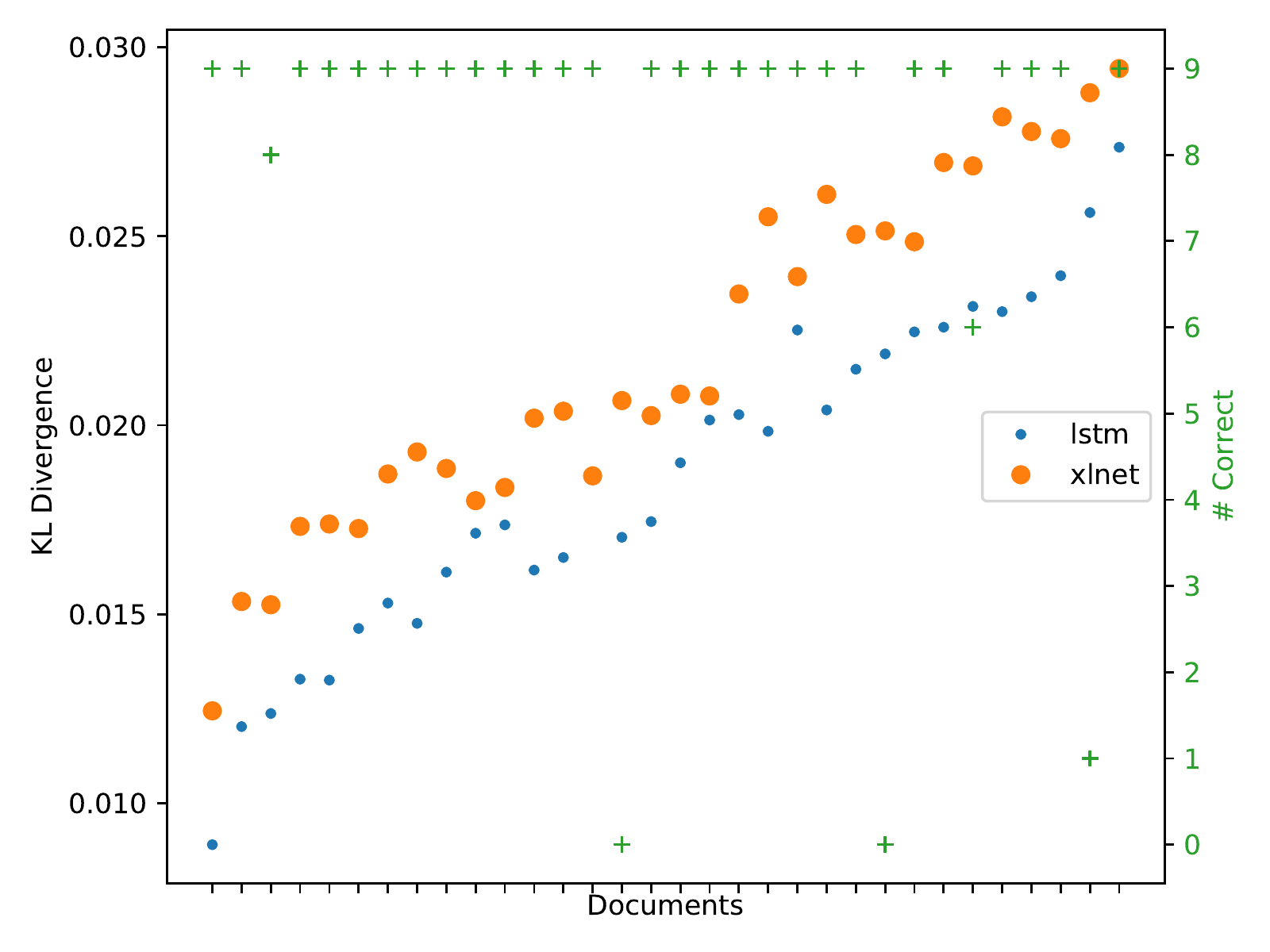}
    \caption{XLNet versus humans --- KL divergence and number of correct models per document.}
    \label{fig:XLNET_KL_Correctness}
  \end{subfigure}
  \caption{Models attention vs. human visual attention. On the x-axis we show each of the 32 documents with the corresponding KL divergence score on the left y-axis. We plot performance of LSTM (cf. Figure~\ref{fig:LSTM_KL_Correctness}) and XLNet (cf. Figure~\ref{fig:XLNET_KL_Correctness}) models for each document with green plus signs as the number of correct models indicated on the right y-axis. In Figure~\ref{fig:LSTM_KL_Correctness}, the larger blue dots show the LSTM divergence score for each document, while the smaller orange dots show the divergence score of XLNet models. Vice-Versa, in Figure~\ref{fig:XLNET_KL_Correctness}, the larger orange dots show the XLNet score for each document, while the smaller blue dots show the divergence score of the LSTM models. The documents are ordered by ascending divergence score.}
\end{figure*}

In order to explore the relationship of model performance and similarity between model attention and human visual attention, we plot in Figures~\ref{fig:LSTM_KL_Correctness} and~\ref{fig:XLNET_KL_Correctness} the nine best {LSTM and XLNet} models performances for each document, sorted by the sum of divergence scores and number of correct models. Similar comparison between CNN and XLNet models can be found in the Appendix, Figure \ref{fig:CNN_LSTM} and \ref{fig:CNN_XLNET}.
Performance, i.e. correctness, refers to the number of models within the ensemble that provided correct answer. The y-axis represents the KL divergence on the left, while the x-axis represents the documents (32 in total), and the legend indicates which models the datapoints refer to. The documents presented on the left of the figure are part of the easier ones and the divergence scales up as document difficulty increases.
When models are faced with difficult questions, we observe performance drops and this seems to be at a specific KL threshold. We suppose that this behavior aligns with the observations reported in the case study from ~\cite{blohm2018comparing}, where human annotators required several strategies to solve difficult questions.
Moreover, our plots show a correlation between attentive LSTM and CNN model performance and similarity to human visual attention.

\label{sec:res_within}
\begin{table}[h]
  \adjustbox{max width=\columnwidth}{
  \begin{tabular}{r r r r} 
    \toprule
    Nine Best & Val Accuracy & Spearman & p$-$value
    \\ [0.5ex] 
    \midrule
    LSTM & 84.37\% & \textbf{-0.73} & \textbf{$<$ 0.001}
    \\ 
    
    CNN & 82.58\% & \textbf{-0.72} & \textbf{$<$ 0.001}
    \\
    
    XLNet & 91.00\% & -0.16 & $0.381$\\
    \bottomrule
  \end{tabular}
  }
  \caption{Spearman's rank correlation coefficients between the number of models which correctly answered a given question on each document and the KL divergence between models and human visual attention. Bold numbers indicate statistically significant correlation scores, where p-value $<$ 0.001.}
  \label{tab:Within_correcness}
\end{table}

To quantify the correlation between system performance and dissimilarity between model and human visual attention, we report in Table~\ref{tab:Within_correcness} the majority vote ensemble accuracy scores for each of the nine best models, Spearman's rank correlation coefficients between the KL divergence scores and the number of models that correctly answered questions, and the corresponding p-values. 
As observed in Figure~\ref{fig:LSTM_KL_Correctness} (and Figure \ref{fig:CNN_LSTM} in the Appendix), there are two \textbf{statistically significant negative correlations} from the attentive \textbf{LSTM (-0.73) and CNN (-0.72) models}.
These correlation scores indicate that for either LSTM or CNN, as the number of models that correctly answered a question related to a document increases, the KL divergence of these model types to human visual attention decreases. 
We conclude that there is a correlation between task performance and similarity between neural attention when leveraging LSTM or CNN and human visual attention distributions.

However in contrast, behavior from XLNet models show weak negative correlation of -0.16 and p = 0.381 (cf. Table \ref{tab:Within_correcness}, cf. Figure \ref{fig:XLNET_KL_Correctness}). 
Most XLNet models correctly answer the questions, although the KL divergence increases (cf. Figure \ref{fig:XLNET_KL_Correctness}), i.e. there is no significant correlation between performance and similarity to human visual attention. All the nine XLNet models always provide correct answers. One potential reason could be that we chose documents that are difficult to answer based on an analysis of CNN and LSTM models.

\subsection{Models vs. Models}

\begin{table*}[t]
  \centering
  \adjustbox{max width=\textwidth}{
  \begin{tabular}{r r c r r r r} 
    \toprule
     Nine Best & Avg KL & Combo & Estimate & Std. Error & t-value & p-value
    \\ [0.5ex] 
    \midrule
     \textbf{LSTM} & 0.018 & LSTM vs.\ XLNet & -0.003 & 0.001 & -2.835 & \textbf{$<$ 0.01}
    \\ 
     CNN & 0.020 & LSTM vs.\ CNN  & -0.001 & 0.001 & -1.098 & $0.27$

    \\
    XLNet & 0.022 & CNN vs.\ XLNet & -0.001 & 0.001 & -1.736 & $0.17$
    
    \\
    \bottomrule
  \end{tabular}
  }
  \caption{Pairwise comparison of the average KL divergence for the three models. Here we show the comparison of each model against each other (LSTM vs. CNN, LSTM vs. XLNET, and CNN vs. XLNet). We compare the models to show if the differences in attention distributions between models is of statistical significance; the significantly different model type (LSTM) can be seen in bold, where p-value $<$ 0.01.}
  \label{tab:Across_Turkys}
\end{table*}

In Table~\ref{tab:Across_Turkys}, we perform a pairwise comparison of the average KL divergence for the three neural models using a linear regression model with Tukey's alpha adjustment method~\cite{sinclair2013alpha}. 
Interestingly, there is a \textbf{statistically significant} difference between the KL divergence of \textbf{LSTMs compared to XLNets} ($\beta = -0.003, p < 0.01$). Even though the performance of the XLNets are better with respect to accuracy, LSTMs are significantly more similar to human visual attention.

This observation suggests that even though aiming to interpret the black box by comparing it to human performance provides insight, we should not force all model types to emulate human visual attention while performing the same task. 

%%%%%%%%%%%%%%%%%%%%%%%%%%%%%%%%%%%%%%%%%%%%%%%%%%%%%%%%%%%%%%%%%%%

\section{Conclusion and Future Work}

Our core contribution is a new method for comparing human visual attention versus neural attention distributions in machine reading comprehension. To the best of our knowledge, we are the first to do so with gaze data. Our findings show that CNNs and LSTMs have a statistically significant correlation between similarity to human visual attention distributions and system performance.
Interestingly, the same is not true for XLNets. Moreover, the attention weights of the LSTMs are significantly different compared to the XLNets. 
Although these pre-trained Transformer networks are less similar to human visual attention, our fine-tuned model obtains the new SOTA on the MovieQA benchmark dataset with 91\% accuracy on the validation set. In addition, we extend the MovieQA dataset with eye tracking data, release this as open source and present an attentive reading visualization tool that supports users to gain insights when comparing human versus neural attention.

In future work we plan to extend our understanding of these large-scale pre-trained language models.
It would be interesting to investigate whether the observed increase in performance but lack of similarity to humans in the XLNet models is because they are pre-trained on large external corpora or whether this is due to inherent properties in architecture, when compared to other pre-trained models (such as BERT). Lastly, to further disentangle token level saliency versus cognitive load of processing, additional analyses and metrics could be considered.

\section{Acknowledgements}

E. Sood was funded by the Deutsche Forschungsgemeinschaft (DFG, German Research Foundation) under Germany's Excellence Strategy - EXC 2075 -- 390740016;
A. Bulling was funded by the European Research Council (ERC; grant agreement 801708);
S. Tannert was supported by IBM Research AI through the IBM AI Horizons Network;
N.T. Vu was funded by the Carl Zeiss Foundation.
We would like to especially thank Manuel Mager for his valuable feedback and guidance. And to Pavel Denisov and Sean Papay for their helpful insights and suggestions.
We would also like to thank Glorianna Jagfeld for her contributions on the dataset, and Fabian K\"ogel for his contributions on the visualization tool.
Lastly, we would like to thank the anonymous reviewers for their useful feedback.

%%%%%%%%%%%%%%%%%%%%%%%%%%%%%%%%%%%%%%%%%%%%%%%%%%%%%%%%%%%%%%%%%%%

\bibliography{references}
\bibliographystyle{acl_natbib}

\appendix

\section{Appendix}

\subsection{Analysis Results -- Models vs. Humans}
\begin{figure}[!th]
  \centering
  \begin{subfigure}{.45\textwidth}
    \centering
    \includegraphics[width=\linewidth]{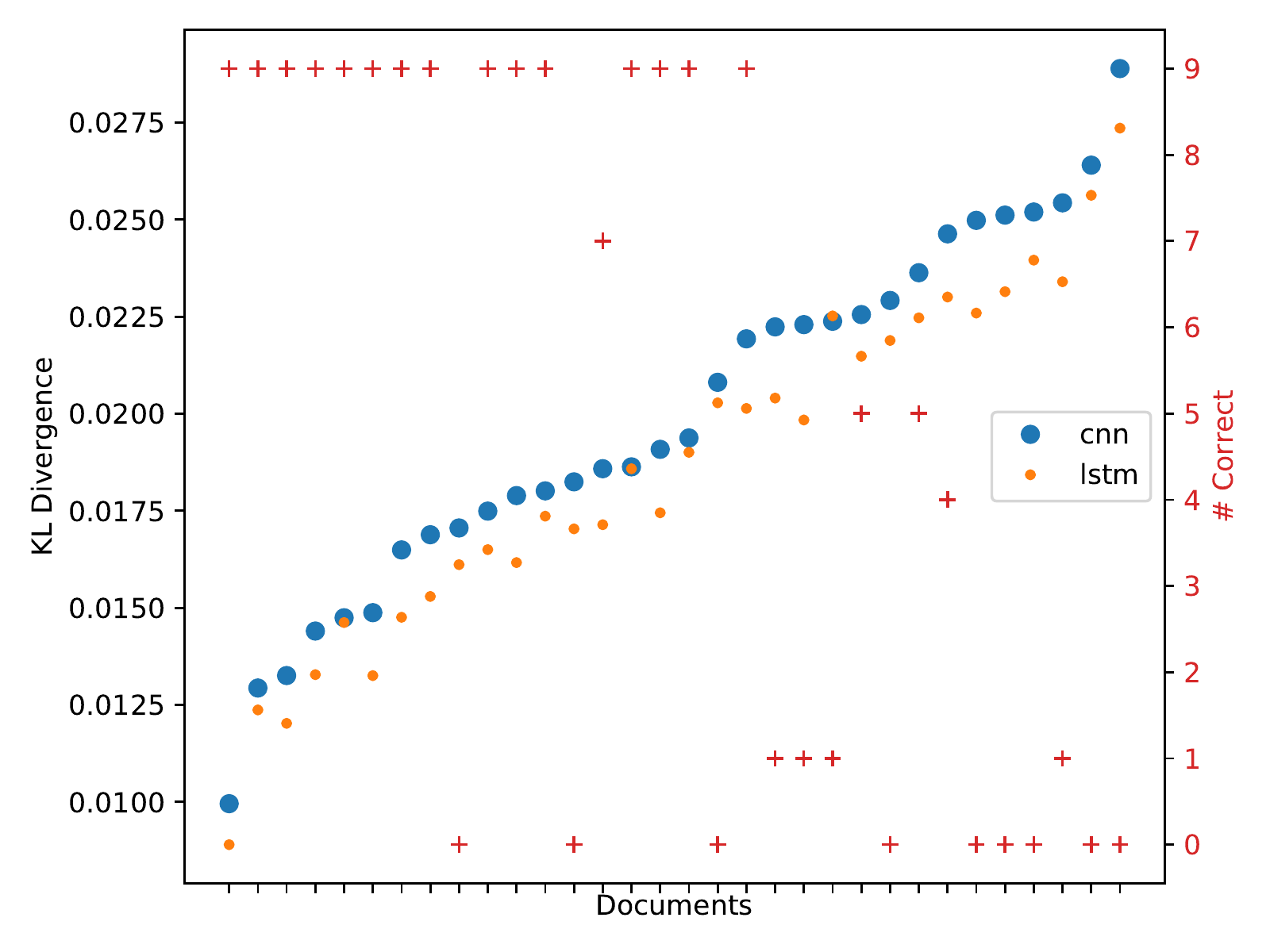}
    \caption{CNN and LSTM versus humans --- KL divergence and number of correct CNN models per document.}
     \label{fig:CNN_LSTM}
  \end{subfigure}
  \hfill
  \begin{subfigure}{.45\textwidth}
    \centering
    \includegraphics[width=\linewidth]{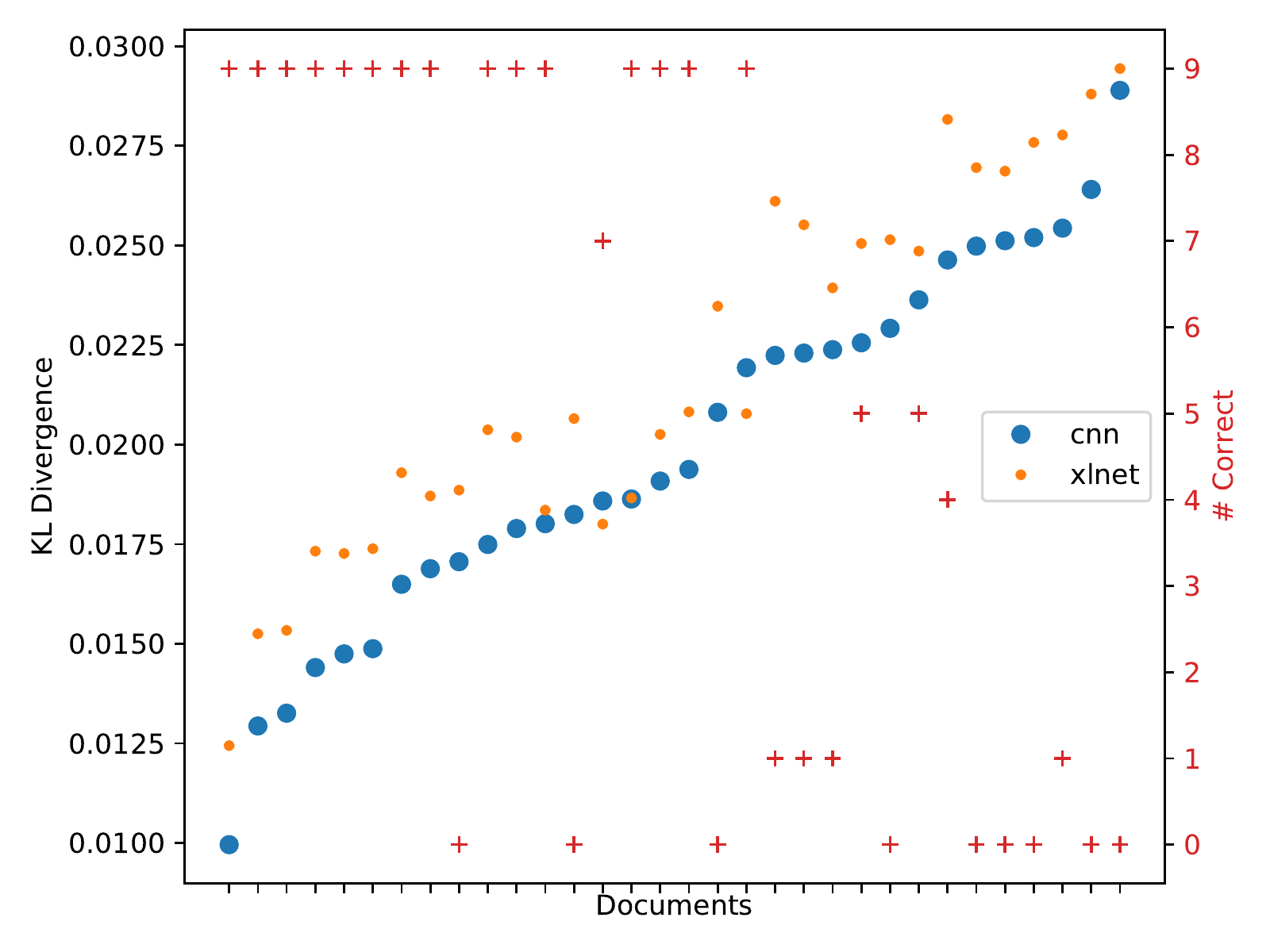}
    \caption{CNN and XLNet versus humans --- KL divergence and number of correct CNN models per document.}
     \label{fig:CNN_XLNET}
  \end{subfigure}
  \caption{In this Figure we show the KL divergence to human attention of CNN and the LSTM models (cf. Figure~\ref{fig:CNN_LSTM} as well as of CNN and the XLNet models (cf. Figure~\ref{fig:CNN_XLNET}) to point out the differences between models. The CNN model divergences are highlighted in the large blue dots, and LSTM and XLNet models are indicated in smaller orange dots. The correctness (in red) indicated on the right y-axis, shows the number correct CNN models per document.}
\end{figure}

\subsection{Visualization Tool}
\begin{figure}[!ht]
  \centering
  \begin{subfigure}{.45\textwidth}
    \centering
    \includegraphics[width=\linewidth]{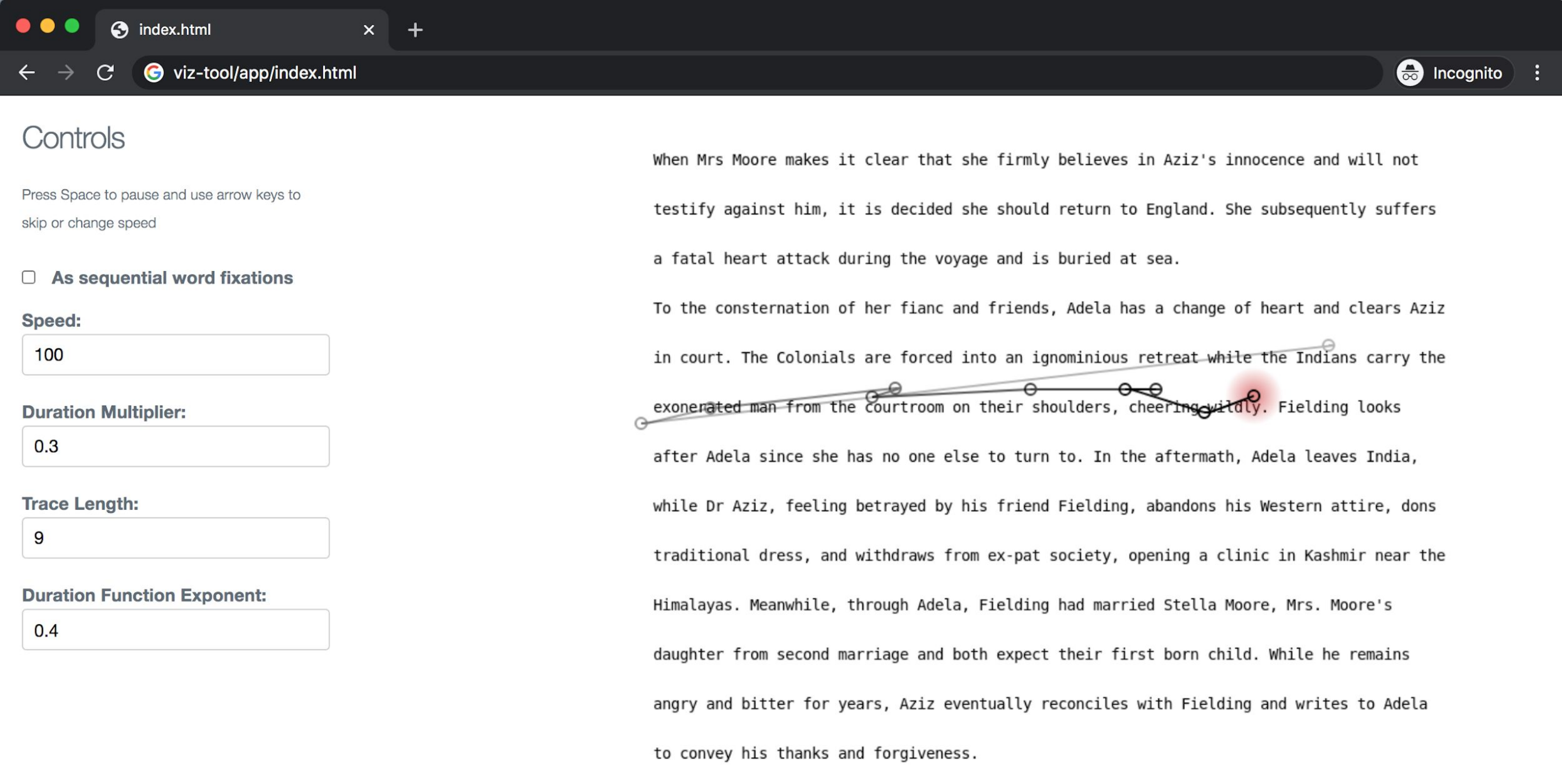}
    \caption{Interface for visualization tool and example of visualization scan path.}
     \label{fig:interface}
  \end{subfigure}
  \hfill
  \begin{subfigure}{.3\textwidth}
    \centering
    \includegraphics[width=\linewidth]{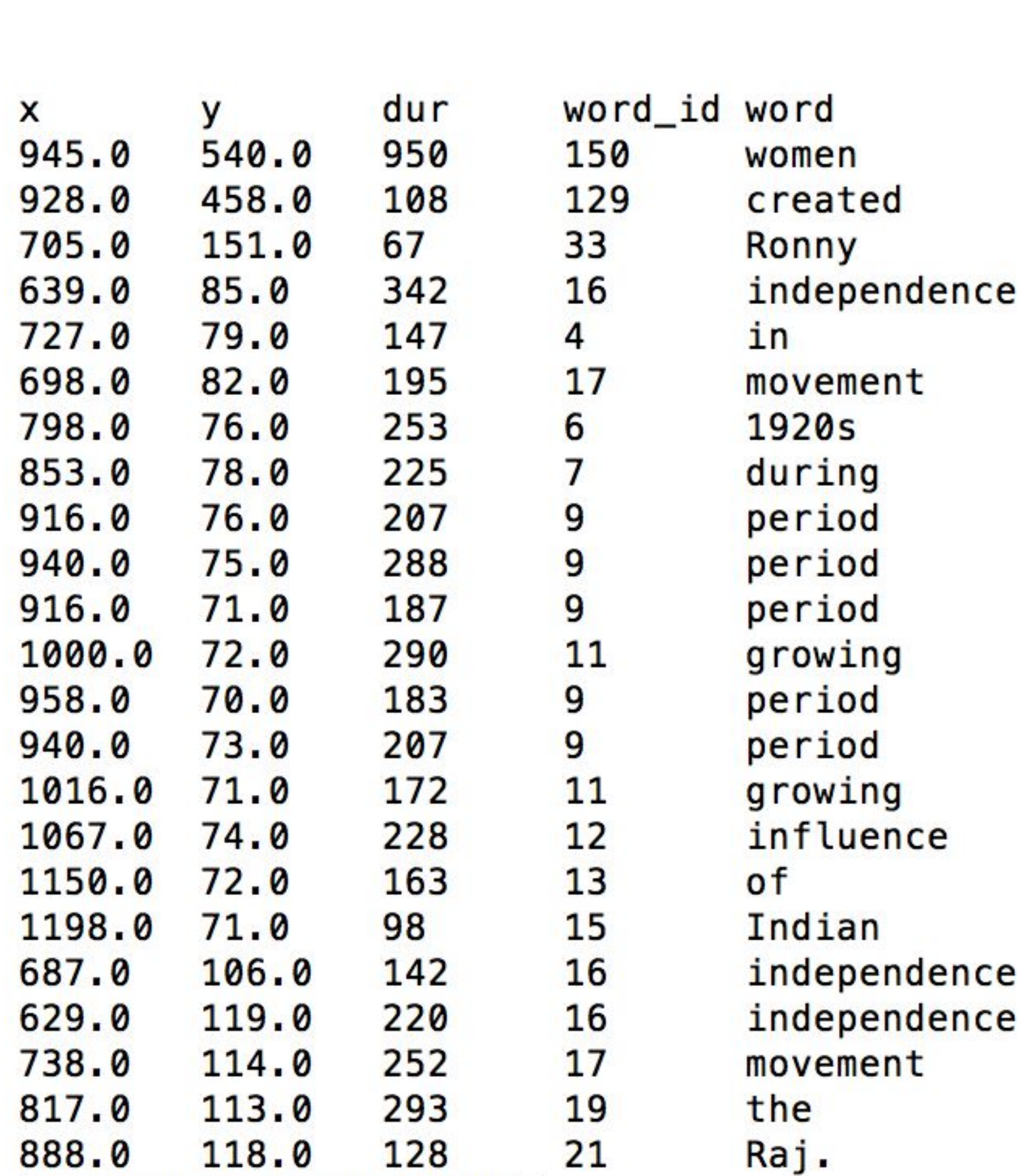}
    \caption{Eye tracking data file required for visualization tool}
     \label{fig:example_data}
  \end{subfigure}
  \caption{Figure~\ref{fig:interface} shows the control options (left side) that allow users to pause the visualization with the space bar, change the speed, duration variables, and length of the scan path. Figure~\ref{fig:interface}, on the right side, shows an example txt stimuli file and the simulated scan path. The red dot indicates fixation duration and expands given the duration length (what we extract as human attention weights). In Figure~\ref{fig:example_data} we show an example of the gaze data txt file required for visualization tool.} 
\end{figure}
 
 \subsection{Coreference Resolution}
 \begin{figure}[!ht]
  \centering
  \includegraphics[width=.55\linewidth]{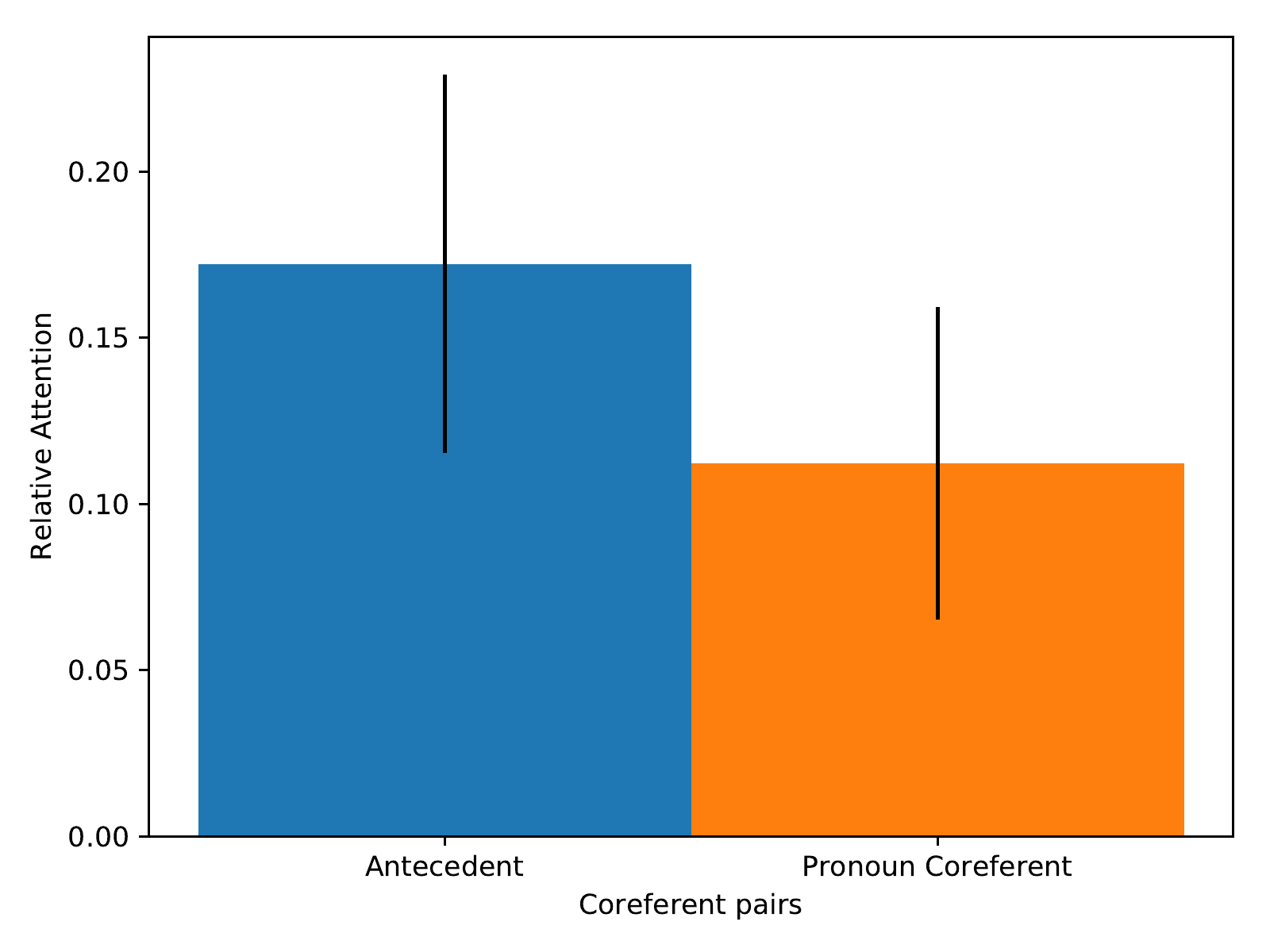}
 \caption{Here we compute the relative importance of coreference chains observed in the human data, where we use fixation durations to denote saliency. We show the agreement in our MQA-RC dataset, between humans, that antecedents are more salient compared to pronoun co-reference chains.}
 \label{fig:saliency_coref}
 \end{figure}
\noindent \textbf{The coreference annotation} 
We used the off-the-shelf high performing coreference model~\cite{lee2018higher} (following the implementation from  \url{https://github.com/kentonl/e2e-coref}), in order to obtain coreference chains over our MQA-RC dataset. We train the model on the same data as reported in ~\cite{lee2018higher}, reproducing reported results over the OntoNotes data, that is, the CoNLL 2012 version of it; thus the model predictions are based on the annotation schema defined in~\cite{pradhan2012conll}. We then test the model on the MQA-RC dataset to obtain our coreference chains. We prepared the data with the automatically generated coreference chain predictions (antecedents and their corresponding pronouns coreference chains), into a web-based annotation tool, WebAnno3~\cite{eckart-de-castilho-etal-2016-web}. At this point, two experienced annotators (one English native speaker, and the other near-native) checked and corrected the automatically generated annotations. The annotators obtained 100\% agreement; we suppose this is due to the small amount of documents, short length of sentences in the documents, and the documents contain easy to resolve pronouns (as seen in Figure \ref{fig:saliency_coref}). We then merge the corrected annotations between annotators, and present this as our coreference annotation over the 32 documents.

\subsection{Extracting LSTM and CNN Word Level Attention}
\begin{figure}[!th]
  \centering
  \begin{subfigure}{.48\textwidth}
    \centering
    \includegraphics[width=\linewidth]{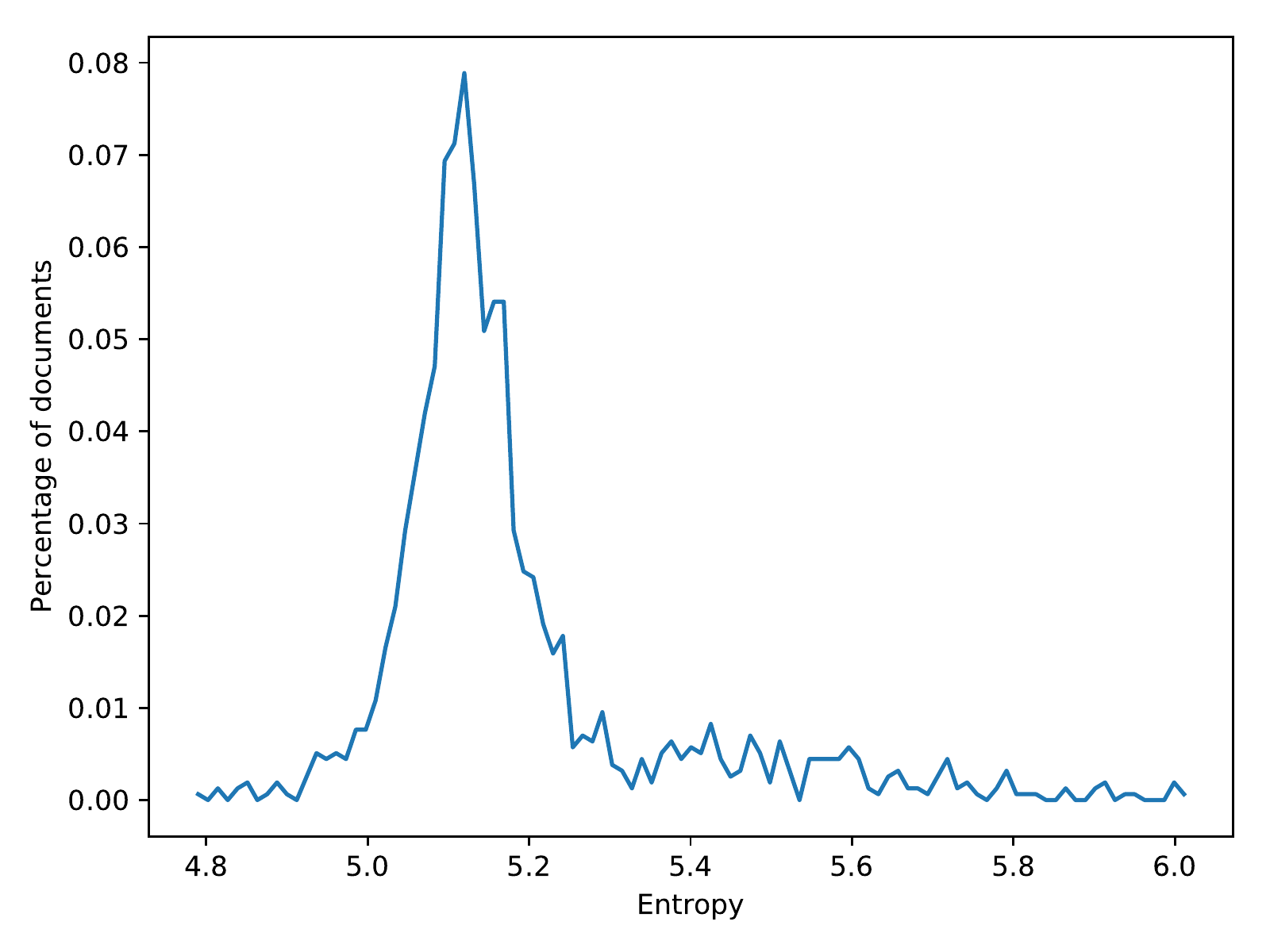}
    \caption{CNN word level attention distribution}
     \label{fig:CNN_entropy_word}
  \end{subfigure}
  \hfill
  \begin{subfigure}{.48\textwidth}
    \centering
    \includegraphics[width=\linewidth]{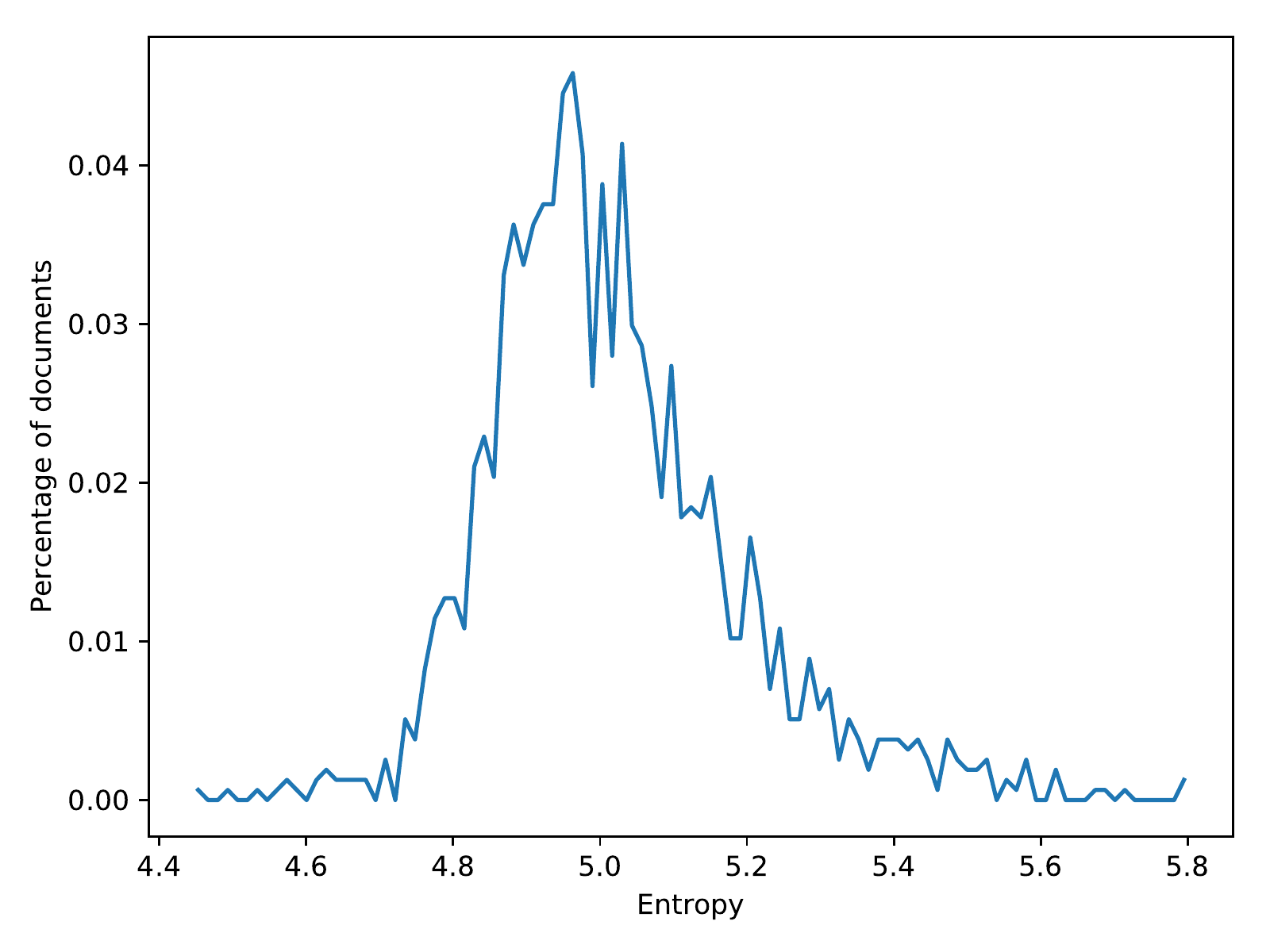}
    \caption{LSTM word level attention distribution}
     \label{fig:LSTM_entropy_word}
  \end{subfigure}
  \caption{We show the word level attention distributions for both CNN~\ref{fig:CNN_entropy_word} and LSTM~\ref{fig:LSTM_entropy_word}. The word level attention distribution has high entropy, and thus provide a suitable option to compare to human attention.}
\end{figure}

% %%%%%%%%%%%%%%%%%%%%%%%%%%%%%%%%%%%%%%%%%%%%%%%%%%%%%%%%%%%%%%%%%%%
\end{document}